\documentclass{article} % For LaTeX2e
\usepackage{iclr2025_conference,times}

\usepackage{wrapfig}
\usepackage{lipsum} % 用于生成示例文字
\usepackage{wrapfig}

\usepackage{amsmath,amsfonts,bm}
\usepackage{multicol}
\def\eqref#1{equation~\ref{#1}}
\def\1{\bm{1}}

\DeclareMathAlphabet{\mathsfit}{\encodingdefault}{\sfdefault}{m}{sl}
\SetMathAlphabet{\mathsfit}{bold}{\encodingdefault}{\sfdefault}{bx}{n}

\usepackage{hyperref}
\usepackage{url}
\usepackage{multicol}
\usepackage{aliascnt}
\usepackage{adjustbox}
\usepackage{tcolorbox}
\usepackage{booktabs}
\usepackage{multirow} 
\usepackage{enumitem}
\usepackage{CJKutf8} % 支持中文
\definecolor{uclablue}{rgb}{0.15, 0.45, 0.68}
\hypersetup{
    breaklinks,
    citecolor=uclablue,
    colorlinks=true,
}

\title{Chinese SimpleQA: A Chinese Factuality Evaluation for Large Language Models}
\author{Yancheng He$^{*}$, Shilong Li$^{*}$, Jiaheng Liu$^{*\dagger}$, \textbf{Yingshui Tan,} \textbf{Weixun Wang,} \textbf{Hui Huang,} \\ \textbf{Xingyuan Bu,} \textbf{Hangyu Guo,}  \textbf{Chengwei Hu,} \textbf{Boren Zheng,} \textbf{Zhuoran Lin,} \textbf{Xuepeng Liu,} \\ \textbf{Dekai Sun,} \textbf{Shirong Lin,}  \textbf{Zhicheng Zheng,} \textbf{Xiaoyong Zhu,} \textbf{Wenbo Su,} \textbf{Bo Zheng}\\
Taobao \& Tmall Group of Alibaba
}

\begin{document}
\maketitle
\let\oldthefootnote\thefootnote

\let\thefootnote\relax\footnotetext{* First three authors contributed equally. ~~$^\dagger$ Corresponding Author: Jiaheng Liu.}
\let\thefootnote\oldthefootnote
\begin{abstract}
% \jh{Need Modify then} 
New LLM evaluation benchmarks are important to align with the rapid development of Large Language Models (LLMs).
In this work,
we present \textbf{Chinese SimpleQA}, the first comprehensive Chinese benchmark to evaluate the factuality ability of language models to answer short questions, and Chinese SimpleQA mainly has five properties (i.e., Chinese, Diverse, High-quality, Static, Easy-to-evaluate).
Specifically,
first, 
we focus on the \textbf{Chinese} language over 6 major topics with 99 \textbf{diverse} subtopics.
Second, we conduct a comprehensive quality control process to achieve \textbf{high-quality} questions and answers,
where the reference answers are \textbf{static} and cannot be changed over time.
Third,
following SimpleQA, the questions and answers are very short, and the grading process is \textbf{easy-to-evaluate} based on OpenAI API.
Based on Chinese SimpleQA, we perform a comprehensive evaluation on the factuality abilities of existing LLMs.
Finally, we hope that Chinese SimpleQA could guide the developers to 
better understand the Chinese factuality abilities of their models and facilitate the growth of foundation models. 
% evaluate the performance of 
% We prioritized two properties in designing this eval. First, SimpleQA is challenging, as it is adversarially collected against GPT-4 responses. Second, responses are easy to grade, because questions are created such that there exists only a single, indisputable answer. Each answer in SimpleQA is graded as either correct, incorrect, or not attempted. A model with ideal behavior would get as many questions correct
% as possible while not attempting the questions for which it is not confident it knows the correct answer. SimpleQA is a simple, targeted evaluation for whether models ``know what they know'' and our hope is that this benchmark will remain relevant for the next few generations of frontier models.
\end{abstract}

\section{Introduction}
A significant challenge in AI development is to ensure language models generate factually accurate responses. 
Current frontier models sometimes produce false outputs or answers that are not substantiated by evidence.
This is the problem known as ``hallucinations'',
which greatly hinders the extensive use of general AI technologies, such as large language models (LLMs). 
Besides,
it is difficult to evaluate the factuality abilities of the existing LLMs. For example, LLMs usually generate lengthy responses containing numerous factual claims. 
Recently, to address the aforementioned evaluation problem, OpenAI has released the SimpleQA benchmark~\citep{Wei2024MeasuringSF} with 4,326 concise and fact-seeking questions, which makes measuring factuality simple and reliable.

% \begin{figure}[!h]
%     \centering
%     \includegraphics[width=\linewidth]{intro.pdf}
%     \caption{Fake Demo Figure.}
%     % \vspace{-4mm}
%     \label{fig:teaser}
% \end{figure} 

However, the SimpleQA benchmark primarily targets the English language, resulting in a limited understanding of LLMs’ capabilities in other languages.
Moreover,
inspired by several recent Chinese LLM benchmarks (e.g., C-Eval~\citep{huang2023ceval}, CMMLU~\citep{li2023cmmlu}),
to evaluate the factuality abilities of LLMs in  Chinese,
we present the \textbf{Chinese SimpleQA} benchmark\footnote{\url{https://openstellarteam.github.io/ChineseSimpleQA/}}, which consists of 3000 high-quality questions spanning 6 major topics, ranging from humanities to science and engineering, as shown in {Figure~\ref{fig: category}}.
Specifically,
% Following SimpleQA,
the distinct main features of our proposed  Chinese SimpleQA dataset are as follows:
\begin{figure}[t]
\centering
\includegraphics[scale=0.9]{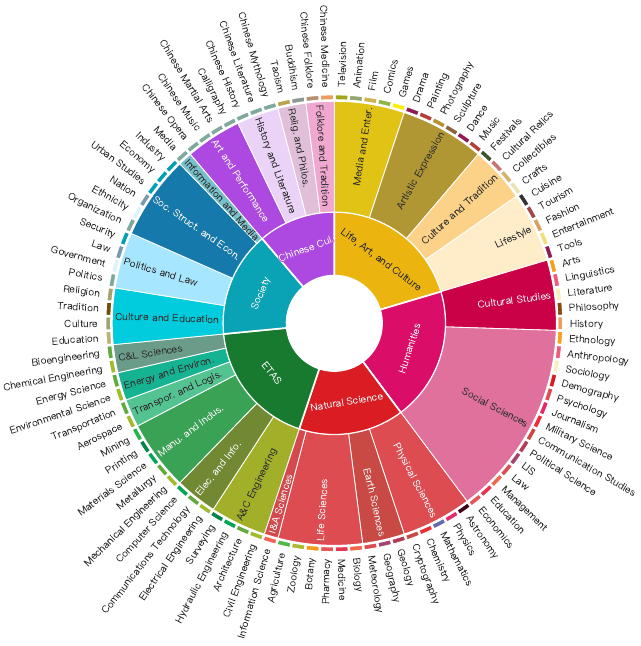} % Adjust the scale as needed
% \vspace{-0.6cm}
\caption{Overview of Chinese SimpleQA. ``Chinese Cul.'' and ``ETAS'' represent ``Chinese Culture'' and ``Engineering, Technology, and Applied Sciences'', respectively.}
\label{fig: category}
% \vspace{-0.6cm}
\end{figure}

\begin{itemize}[leftmargin=8mm]
    \item \textbf{Chinese}: Our Chinese SimpleQA focuses on the Chinese language, which provides a comprehensive evaluation of the factuality abilities of existing LLMs in Chinese.
    % \item In our DDK, we first propose to construct the reweighting factor based on the performance discrepancy across different domains and introduce the high-order factor updating mechanism
 % to improve the stability of the DDK method.
    \item \textbf{Diverse}: Chinese SimpleQA covers 6 topics (i.e., ``Chinese Culture'', ``Humanities'', ``Engineering, Technology, and Applied Sciences'', ``Life, Art, and Culture'', ``Society'', and ``Natural Science''), and these topic includes 99 fine-grained subtopics in total, which demonstrates the diversity of our Chinese SimpleQA.
    % and employs a meticulous loss function that balances the intrinsic language modeling capabilities of the student LLM with the external distillation influences derived from the teacher model.
        \item \textbf{High-quality}: We conduct a comprehensive and rigorous quality control process to ensure the quality and accuracy of our Chinese SimpleQA.
        % \jh{Give more important details}
    \item \textbf{Static}: Following SimpleQA, to preserve the evergreen property of Chinese SimpleQA, all reference answers would not change over time.

    \item \textbf{Easy-to-evaluate}: Following SimpleQA, as the questions and answers are very short, the grading procedure is fast to run via existing LLMs (e.g., OpenAI API).

    % , making the student LLMs perform competitively to the 2x size counterparts.
\end{itemize}
Moreover,
% we perform co
% Based ,
we  perform a comprehensive evaluation and analysis of existing LLMs on Chinese SimpleQA, and several insightful findings are as follows:

\begin{itemize}[leftmargin=8mm]
    \item \textbf{Chinese SimpleQA is challenging}. Only o1-preview and Doubao-pro-32k achieve the passing score (63.8\% and 61.9\% on the correct metric), and there is a long way to improve for many closed-source and open-source LLMs. 
    
    \item \textbf{Larger models lead to better results}. Based on the results of the Qwen2.5 series, InternLM series, Yi-1.5 series, etc, we observe that better performance is obtained when the model is larger.
    
    \item \textbf{Larger models are more calibrated}. 
We observe that o1-preview is more calibrated than o1-mini, and GPT-4o is more calibrated than GPT-4o-mini.

    \item \textbf{RAG matters}. When introducing the RAG strategy into existing LLMs, the performance gaps between different LLMs decrease a lot. For example, for GPT-4o and Qwen2.5-3B, the performance gap decreases from 42.4\% to 9.3\%  within using RAG.
    % \item COT

    \item \textbf{Alignment tax exists}. Existing alignment or post-training strategies usually decrease the factuality of language models.
    \item \textbf{Rankings of SimpleQA and Chinese SimpleQA are different}.
    The performance of several LLMs focusing on Chinese (Doubao-pro-32k, and GLM-4-Plus) is close to the high-performance o1-preview. In particular, in the ``Chinese Culture'' topic, these Chinese community LLMs are significantly better than GPT or o1 series models.
    % Chinese community LLMs achieve better performance on Chinese SimpleQA
    % , making the student LLMs perform competitively to the 2x size counterparts.
\end{itemize}

% first, 
% to 
% where the SimpleQA a simple and reliable dataset for measuring the factuality of frontier models.
% To address this, our research narrows the focus to brief, fact-based questions with singular answers. This approach enhances the feasibility of measuring factuality, though it leaves certain research questions unresolved, such as whether improvements in short-form accuracy translate to long-form content.

% We introduce SimpleQA, a benchmark comprising 4,326 concise, fact-oriented questions. This dataset was developed with several key features in mind:

% Accuracy: Answers are verified by two independent AI experts, with questions formulated for easy evaluation.

% User-friendly: The brevity of questions and answers allows for quick implementation and grading via AI APIs, while the sample size ensures statistical reliability.

% Challenging: Unlike older, saturated benchmarks, SimpleQA is designed to test the limits of cutting-edge models, with top performers currently achieving less than 50% accuracy.

% Comprehensive: The questions span various subjects, including history, science, technology, arts, geography, and popular culture.

% SimpleQA aims to provide a straightforward, dependable method for assessing the factual accuracy of advanced language models.

\section{Chinese SimpleQA}

\subsection{Overview}
% 表格和现有数据的diff（simpleQA, WebQA, NQA），打对号
% ``Correct given attempted''

Figure \ref{fig: category} shows the category distribution of Chinese SimpleQA, 
which contains six primary topics: ``Chinese Culture'', ``Humanities'', ``Engineering, Technology and Applied Sciences'', ``Life, Art and Culture'', ``Society'', and ``Natural Science''. Each primary topic includes multiple secondary subtopics. 
% covering nearly all fields of knowledge in Chinese. T
In Table \ref{tab: benchmark_compare}, we also compare Chinese SimpleQA with several mainstream LLMs evaluation benchmarks,
which demonstrates that Chinese SimpleQA is the first benchmark to focus on the evaluation of the boundaries of Chinese knowledge in LLMs.

\begin{table}[!t]
    \centering
    \small
    \resizebox{\textwidth}{!}{
    \begin{tabular}{lcccccc}
        \toprule
        % \textbf{Benchmark} & \multicolumn{4}{c}{\textbf{Dataset Information}} & \multicolumn{3}{c}{\textbf{Evaluation Method}} \\
        % \cmidrule(lr){2-5} \cmidrule(lr){6-8}
         \textbf{Benchmark} & \textbf{Data Size} & \textbf{Language} & \textbf{Data Source} & \textbf{Domain} & \textbf{Reasoning} & \textbf{Metric} \\
        \midrule
        WebQA~\citep{li2016dataset} & 42187 & Chinese & Real World & Knowledge & \checkmark & Accuracy \\
        MMLU~\citep{mmlu} & 15,908 & English & Exams \& Textbooks & Knowledge &\checkmark & Accuracy \\
        CMMLU~\citep{Li2023CMMLUMM} & 11,528 & Chinese & Exams & Knowledge & \checkmark & Accuracy \\
        GSM8K~\citep{math-verifier} & 8,792 & English & Human Writers & Math &  \checkmark & Accuracy \\
        AlpacaEval~\citep{alpaca_eval} & 805 & English & Alpaca Data & General &  \checkmark & LLM-as-a-Judge \\
        MT-Bench~\citep{llm-as-a-judge} & 80 & English & Self-constructed & General &  \checkmark & LLM-as-a-Judge \\
        Arena-Hard~\citep{Li2024FromCD} & 500 & English & Human Writers & General &  \checkmark & LLM-as-a-Judge \\
        % HumanEval~\citep{codex} & 164 & Python & Human Writers & Code & \checkmark & \texttimes & Pass@k \\
        % AGI-Eval (Zhong et al., 2023) & 8,062 & Chi. \& Eng. & Exams & Knowledge & \texttimes & \texttimes & Accuracy \\
        C-Eval~\citep{huang2023ceval} & 13,948 & Chinese & Exams & Knowledge &  \checkmark & Accuracy \\
        SimpleQA~\citep{Wei2024MeasuringSF} & 4,326 & English & Human Writers & Knowledge  & \texttimes & LLM-as-a-Judge \\
        \midrule
        \textbf{Chinese SimpleQA (Ours)} & 3000 & Chinese &\begin{tabular}[c]{@{}c@{}}{Self-constructed}\\ {\&Human Writers}\end{tabular}  & Knowledge &  \texttimes & LLM-as-a-Judge \\
        \bottomrule
    \end{tabular}}
    \caption{Comparisons between our Chinese SimpleQA and other benchmarks.}
    \label{tab: benchmark_compare}
\end{table}

\subsection{Data collection}
\label{Data collection}

\begin{figure}[!t]
\centering
% \setlength{\abovecaptionskip}{0.1cm}
% \setlength{\belowcaptionskip}{-0.3cm}  
% \resizebox{0.48\textwidth}{!}{}
\includegraphics[width=1\linewidth]{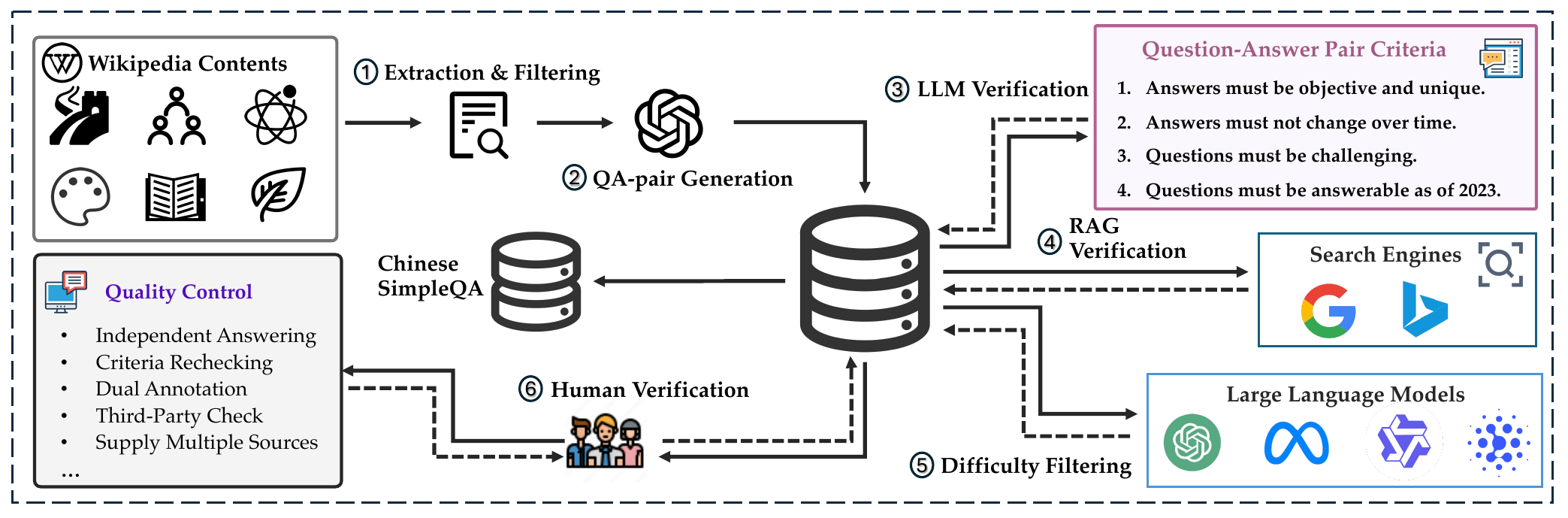}
% \vspace{-0.6cm}
\caption{An overview of the data construction process of Chinese SimpleQA.}
\label{fig: construct}
% \vspace{-0.6cm}
\end{figure}

% Wiki, GPT query&response, 不同模型生成回复，控制query的难度，然后GPT-4校验
% The automated data collection for Chinese SimpleQA comprises question-answer pair generation, condition-based automated validation, and retrieval-based answer accuracy verification. 
As shown in Figure~\ref{fig: construct}, 
the data collection process for Chinese SimpleQA involves both automated construction and human verification. The automated phase includes: (1) extracting and filtering relevant knowledge content, (2) generating question-answer pairs automatically, (3) verifying these pairs using an LLM based on predefined criteria, (4) performing Retrieval-Augmented Generation (RAG) verification, and (5) filtering for difficulty level.

% Initially, we aggregated a substantial amount of knowledge content from diverse fields, which primarily originated from Wikipedia. This content was then processed through a quality assessment model to filter out low-quality data. Based on this, we guided the LLM in generating question-answer pairs according to predefined criteria. To ensure compliance with these criteria, a rule-based validation was conducted with the LLM to eliminate non-conforming data. Relying on a single data source can potentially yield inaccurate answers; thus, to mitigate this risk, we deployed an external retriever to gather more varied information, guiding the LLM in evaluating the accuracy of answers based on information from different sources. Incorrect question-answer pairs were subsequently discarded. Specifically, LlamaIndex was utilized as the retrieval method, with search results from Google and Bing serving as data sources. The detailed generation and validation prompts are provided in Appendix X.
Specifically,
first, we collect a large amount of knowledge-rich text content from various knowledge fields (e.g., Wikipedia),
and we utilize a quality assessment model to filter out low-quality data. Then, we prompt the LLM to generate question-answer pairs using these high-quality knowledge contents. After that, to ensure the quality of Chinese SimpleQA, we use LLM to remove samples, which cannot meet the requirements of our predefined  criteria. 
In this way, we can obtain a large set of initially filtered knowledge question-answer pairs. 
Meanwhile, to improve the quality of answers,
% However, relying on a single data source for generation can potentially lead to inaccurate answers. To mitigate this risk,
we deploy external retrieval tools (i.e., search engines) to gather more diverse information, which guides the LLM to evaluate the factual correctness of answers based on the RAG system. 
% In this process, incorrect question-answer pairs were discarded. 
Specifically, we apply LlamaIndex~\footnote{\url{https://github.com/run-llama/llama_index}} as the retrieval method, with search results from Google and Bing as data sources.
% further enhancing the quality of the dataset.
Details on the generation and validation can be found in Appendix \ref{app: gereration_and_validation}.
% We evaluate 17 closed-source LLMs (i.e., o1-preview~\footnote{\url{https://openai.com/index/introducing-openai-o1-preview/}}, Doubao-pro-32k\footnote{\url{https://www.volcengine.com/product/doubao}}, GLM-4-Plus\footnote{\url{https://bigmodel.cn/dev/api/normal-model/glm-4}}, GPT-4o\footnote{\url{https://openai.com/index/hello-gpt-4o/}}, Qwen-Max~\citep{qwen1.5}, Gemini-1.5-pro~\citep{geminiteam2024gemini15unlockingmultimodal}, 
% DeepSeek-V2.5~\citep{deepseekv2}, 
% Claude-3.5-Sonnet~\footnote{\url{https://www.anthropic.com/news/claude-3-5-sonnet}},
% Yi-Large\footnote{\url{https://platform.lingyiwanwu.com/}}, moonshot-v1-8k\footnote{\url{https://platform.moonshot.cn/}}, GPT-4-turbo~\citep{gpt4}, GPT-4~\citep{gpt4}, Baichuan3-turbo\footnote{\url{https://platform.baichuan-ai.com/}}, o1-mini\footnote{\url{https://openai.com/index/openai-o1-mini-advancing-cost-efficient-reasoning/}}, Doubao-lite-4k\footnote{\url{https://www.volcengine.com/product/doubao}}, GPT-4o-mini\footnote{\url{https://openai.com/index/gpt-4o-mini-advancing-cost-efficient-intelligence/}}, GPT-3.5~\citep{gpt-3},
% and 24 open-source LLMs (i.e., Qwen2.5 series~\citep{qwen2.5}, InternLM2.5 series~\citep{cai2024internlm2}, Yi-1.5 series~\citep{ai2024yiopenfoundationmodels}, LLaMA3~\citep{llama3modelcard} series, DeepSeek Series~\cite{deepseek-llm}, Baichuan2 series~\citep{baichuan2023baichuan2}, Mistral series~\citep{jiang2023mistral}, ChatGLM3 and GLM-4~\citep{glm2024chatglm, du2022glm}).
In addition, we filter some simple samples to discover the knowledge boundaries of the LLMs and improve the difficulty of Chinese SimpleQA. Specifically, if a question could be correctly answered by all four powerful models\footnote{GPT-4o~\citep{gpt4}, Meta-Llama-3-70B-Instruct~\citep{dubey2024llama3}, Qwen2.5-72B-Instruct~\citep{qwen2.5}, and GLM-4-Plus\footnote{\url{https://bigmodel.cn/dev/api/normal-model/glm-4}}.}, it is considered as a simple question and will be discarded. 
% Through this approach, Chinese SimpleQA becomes more challenging.

Notably, the construction of question-answer pairs is based on the following criteria:

\paragraph{Answers must be objective and unique.}
Questions should relate to factual knowledge about the objective world and remain uninfluenced by personal subjective views. For example, questions beginning with ``What do you think about'' or ``How would you evaluate'' are inappropriate. Furthermore, the answer to each question must be unique, precluding the possibility of multiple correct responses. For instance, the query ``In what year did Zhu Qizhen ascend to the throne?'' is inadequate because it has two possible answers: 1435 and 1457.
\paragraph{Answers must not change over time.}
Answers should consistently reflect timeless facts, unaffected by the time the question is posed. For example, ``What is the atomic number of carbon?'', and the answer ``6" remains unchanged. In contrast, questions regarding current affairs, such as  ``Who is the current president of a certain country?'' are inappropriate, as their answers are subject to change.
\paragraph{Questions must be challenging.}
Questions should not be overly simplistic, and the designed queries need to thoroughly assess the model's depth of knowledge. 
\paragraph{Questions must be answerable as of 2023.}
Each question must be answerable by December 31, 2023, ensuring fair evaluation for models trained on data available post this date.

\subsection{Quality Control}
% Human annotation规则，(Guideline), 过滤掉xx%的数据
Following automated data collection, we employ human verification to enhance dataset quality. Specifically, each question is independently assessed by two human annotators. Initially, annotators determine whether the question adheres to the aforementioned predefined criteria. If either annotator deems the question non-compliant, this sample is discarded. Subsequently, both annotators utilize search engines to retrieve pertinent information and formulate answers. During this stage, the annotators are supposed to use content from authoritative sources (e.g., Wikipedia\footnote{\url{https://www.wikipedia.org/}}, Baidu Baike\footnote{\url{https://baike.baidu.com/}}), and each annotator must provide at least two supporting URLs. In cases where the annotators' answers are inconsistent, a third annotator reviews the sample. The final annotation is determined by the third annotator, referencing the initial two assessments. Finally, the human annotation results are compared with responses generated by the LLM, retaining only question-answer pairs that are entirely consistent. This rigorous human verification process ensures that our dataset maintains high accuracy and meets established standards.

In the entire process of constructing and annotating Chinese SimpleQA, many low-quality question-answer pairs are discarded.
% filtered to retain only high-quality ones. 
Specifically, 10,000 pairs are initially generated. After difficulty evaluation through testing with different models, roughly 6,310 pairs are retained, with about 37\% of the easier data being discarded. Following this, another 2,840 samples are removed after rule-based validation and model-based RAG validation, which means that only about 35\% of the original generated data remains. Finally, after a thorough and rigorous manual review, only about 3,000 samples are kept, which is approximately 30\% of the original dataset. 
% Overall, based on Through different methods of quality filtering at each stage, we ultimately obtained this high-quality evaluation set.

\subsection{Dataset statistics}
% 饼状图，表格，类别体系，每个大类别多少条样本，query seq length柱状图
% Using ChatGPT to classify types of answers, we found that 32.8% of answers were dates, 24.1% of answers were a person, 15.3% answers were a number, 9.9% answers were
% a place, and 18.0% answers were classified as “other.”
Table \ref{tab: detail_data} presents the statistics of Chinese SimpleQA. With a total of 3000 samples, the data distribution across the six primary topics in Chinese SimpleQA is relatively balanced, which can effectively assess the knowledge boundaries of LLMs in various fields. Furthermore, the length distribution of both questions and reference answers in this dataset is very short, which is characteristic of knowledge-based queries. 
Notably, evaluating models using Chinese SimpleQA requires minimal input and output tokens, resulting in very low evaluation computation and time costs.

% \begin{table*}[h!]
% \centering
% \caption{Dataset statistics of Chinese SimpleQA.}
% \footnotesize
% \resizebox{\textwidth}{!}{
% \begin{tabular}{lc|lc}
% \toprule
% \textbf{Statistics} & \textbf{Number} & \textbf{Statistics} & \textbf{Number} \\
% \midrule
% \textbf{\#Problems} & 3000 & \textbf{Length} & \\
% \textbf{Primary Category} & & \textit{Question Length} & \\
% - Chinese Culture & 323 & - \textit{maximum length} & 81 \\
% - Humanities & 623 & - \textit{minimum length} & 8 \\
% - Engineering, Technology & 476 & - \textit{avg length} & 23.6 \\
% \phantom{- } and Applied Sciences & & \textit{Reference Answer Length} & \\
% - Life, Art and Culture & 604 & - \textit{maximum length} & 47 \\
% - Society & 450 & - \textit{minimum length} & 1 \\
% - Natural Science & 529 & - \textit{avg length} & 6.1 \\
% \bottomrule
% \end{tabular}}
% \label{tab: detail_data}
% \end{table*}

\begin{table}[!t]
\centering
\small
\caption{Dataset statistics of Chinese SimpleQA.}
{\scriptsize % 设置更小的字体
\setlength{\tabcolsep}{4pt} % 缩小列间距
\resizebox{0.7\textwidth}{!}{ % 调整缩放比例
    \begin{tabular}{lc|lc}
    \toprule
    \textbf{Statistics} & \textbf{Number} & \textbf{Statistics} & \textbf{Number} \\
    \midrule
    \textbf{\#Problems} & 3000 & \textbf{Length} & \\
    \textbf{Primary Topics} & & \textbf{Question Length} & \\
    - Chinese Culture & 323 & - \textit{maximum length} & 81 \\
    - Humanities & 623 & - \textit{minimum length} & 8 \\
    - Engineering, Technology & 473 & - \textit{avg length} & 23.6 \\
    \phantom{- } and Applied Sciences & & \textbf{Reference Answer Length} & \\
    - Life, Art and Culture & 602 & - \textit{maximum length} & 47 \\
    - Society & 450 & - \textit{minimum length} & 1 \\
    - Natural Science & 529 & - \textit{avg length} & 6.1 \\
    \bottomrule
    \end{tabular}
}
}
\label{tab: detail_data}
\end{table}

\subsection{Evaluation Metrics}
Following SimpleQA, we also adopt the following five evaluation metrics.

\begin{itemize}[leftmargin=8mm]
    \item \textbf{Correct (CO)}: The predicted answer fully includes the reference answer without introducing any contradictory elements.

    \item \textbf{Not attempted (NA)}: The reference answer is not fully given in the predicted answer, and there are no contradictory elements with the reference answer.
        \item \textbf{Incorrect (IN)}: The predicted answer contradicts the reference answer, even if the contradiction is solved.

    \item \textbf{Correct given attempted (CGA)}: The metric is the proportion of accurately answered questions among those attempted questions.

    \item \textbf{F-score}: The metric represents the harmonic mean between correct and correct given attempted.

    % \item Alignment Tax
    % \item ranking relationships between SimpleQA and Chinese SimpleQA
    % , making the student LLMs perform competitively to the 2x size counterparts.
\end{itemize}

% 基于gpt-4评测openai的评测metrics
% 基于正则的metrics
% 排序的一致性
\section{Experiments}

%第四段：实验结果的分析

%第五段：总结

\subsection{Baseline Models}
% 开源（中英模型，llama3）和闭源模型

We evaluate 17 closed-source LLMs (i.e., o1-preview~\footnote{\url{https://openai.com/index/introducing-openai-o1-preview/}}, Doubao-pro-32k\footnote{\url{https://www.volcengine.com/product/doubao}}, GLM-4-Plus\footnote{\url{https://bigmodel.cn/dev/api/normal-model/glm-4}}, GPT-4o\footnote{\url{https://openai.com/index/hello-gpt-4o/}}, Qwen-Max~\citep{qwen1.5}, Gemini-1.5-pro~\citep{geminiteam2024gemini15unlockingmultimodal}, 
DeepSeek-V2.5~\citep{deepseekv2}, 
Claude-3.5-Sonnet~\footnote{\url{https://www.anthropic.com/news/claude-3-5-sonnet}},
Yi-Large\footnote{\url{https://platform.lingyiwanwu.com/}}, moonshot-v1-8k\footnote{\url{https://platform.moonshot.cn/}}, GPT-4-turbo~\citep{gpt4}, GPT-4~\citep{gpt4}, Baichuan3-turbo\footnote{\url{https://platform.baichuan-ai.com/}}, o1-mini\footnote{\url{https://openai.com/o1/}}, Doubao-lite-4k\footnote{\url{https://www.volcengine.com/product/doubao}}, GPT-4o-mini\footnote{\url{https://openai.com/index/gpt-4o-mini-advancing-cost-efficient-intelligence/}}, GPT-3.5~\citep{gpt-3},
and 24 open-source LLMs (i.e., Qwen2.5 series~\citep{qwen2.5}, InternLM2.5 series~\citep{cai2024internlm2}, Yi-1.5 series~\citep{ai2024yiopenfoundationmodels}, LLaMA3~\citep{llama3modelcard} series, DeepSeek Series~\citep{deepseek-llm}, Baichuan2 series~\citep{baichuan2023baichuan2}, Mistral series~\citep{jiang2023mistral}, ChatGLM3 and GLM-4~\citep{glm2024chatglm, du2022glm}).
% , , Qwen-Max~\citep{qwen1.5}, and DeepSeek2.5~\citep{deepseekv2}. 
% LLaMA3~\citep{llama3modelcard} series, Qwen1.5~\citep{qwen1.5} and Qwen2~\citep{yang2024qwen2} series, Mistral~\citep{jiang2023mistral}, ChatGLM3 and GLM-4~\citep{glm2024chatglm, du2022glm, zeng2022glm} series. 

 % The o1 preview is the strongest and the most advanced API models in the Chinese and English communities, such as the bean bun and glm-4 series, which are close to o1 preview and better than gpt4. Even the 4o (2) mini series models (o1 mini, gpt4o mini) have significantly lower knowledge capabilities than their corresponding models o1 preview. The larger the gpt4o (3) model, the stronger the performance. From qwen2.5 and interplm2.5, it can be seen that qwen2.5-72b has the strongest open source and is better than some API models. The smaller the (4) model, the higher the probability of not being attended, such as qwen2.5-1.8b, interplm2.5-1.8b, o1. There is a significant difference in performance among different subtopics in Mini (5). The Chinese community models (Bean Pack, glm-4, qwen max, Deepseek) are significantly better than the GPT/O1 model in Chinese culture, and O1 still has significant advantages in natural and applied sciences

% \input{tables/main_result}
% Chinese Culture, Life&Art, 
\begin{table*}[t]
\centering

\resizebox{1.0\textwidth}{!}{
\begin{tabular}{c|ccccc|cccccc}
\toprule
\multirow{2}{*}{\textbf{Models}} & \multicolumn{5}{c|}{\textbf{Overall results on 5 metrics}}  & \multicolumn{6}{c}{\textbf{F-score on 6 topics}}\\
\cmidrule(lr){2-12} 
 & \textbf{CO}   & \textbf{NA}    & \textbf{IN}     & \textbf{CGA}  & \textbf{F-score}     & \textbf{CC}        & \textbf{HU}  & \textbf{ETAS}     & \textbf{LAC}        & \textbf{SO} &\textbf{NS}   \\ 

\midrule
\multicolumn{12}{c}{\textit{Closed-Source Large Language Models}}\\ 
\midrule
o1-preview   &63.8&12.2&24.0&72.7&{67.9}&45.7&69.8&72.4&65.0&73.5&72.3 \\
Doubao-pro-32k&61.9&10.3&27.8&69.1&65.3&61.8&69.3&69.0&56.1&64.2&70.4 \\
GLM-4-Plus  &58.7&7.4&33.9&63.4&60.9&56.5&64.1&64.9&50.7&66.6&62.8\\ 
GPT-4o   &59.3&1.4&39.3&60.1&59.7&39.4&64.0&65.1&53.3&68.6&62.0 \\
Qwen-Max &54.1&11.3&34.6&61.0&57.4&47.8&59.9&63.5&49.9&61.2&59.3\\
Gemini-1.5-pro&54.4&8.0&37.6&59.1&56.7&41.4&59.1&60.8&52.2&56.3&64.3\\
DeepSeek-V2.5&54.1&5.9&40.0&57.5&55.7&50.4&57.6&58.8&50.1&59.4&56.9\\
Claude-3.5-Sonnet&46.2 & 27.4 & 26.4 & 63.6 & 53.5 & 28.7 & 61.3 & 60.4 & 42.2 & 59.8 & 57.7  \\
Yi-Large &47.3&16.4&36.3&56.6&51.5&41.1&56.5&55.1&41.7&57.6&53.8\\
moonshot-v1-8k&48.7 & 5.4 & 45.9 & 51.5 & 50.1 & 49.8 & 54.1 & 56.8 & 41.4 & 53.0 & 46.6\\
GPT-4-turbo &45.6&14.2&40.2&53.1&49.1&24.2&55.2&58.9&43.9&52.5&50.8\\
GPT-4    &45.4&8.4&46.2&49.6&47.4&25.2&54.0&52.8&41.8&52.8&50.6\\
Baichuan3-turbo&45.2&9.0&45.8&49.6&47.3&32.3&52.5&54.0&35.4&54.6&50.9\\
o1-mini&39.5&20.6&39.9&49.7&44.1&21.3&49.2&55.9&33.8&48.8&46.8  \\
Doubao-lite-4k&36.7 & 31.2 & 32.1 & 53.3 & 43.4 & 40.2 & 44.8 & 51.0 & 31.1 & 41.4 & 50.4 \\
GPT-4o mini &37.6&0.9&61.5&37.9&37.8&19.0&42.4&46.4&31.0&42.2&39.8\\
GPT-3.5 &29.7&2.9&67.4&30.6&30.1&13.3&35.8&35.2&25.6&32.7&31.7\\

\midrule
\multicolumn{12}{c}{\textit{Open-Source Large Language Models}} \\ \midrule
Qwen2.5-72B&48.4&7.1&44.5&52.1&50.2&36.3&56.1&57.9&37.1&53.3&56.4\\
Qwen2.5-32B&38.8&11.1&50.1&43.6&41.1&33.7&45.8&48.7&27.3&44.7&44.9\\
Qwen2.5-14B&35.4&9.6&55.0&39.2&37.2&30.2&41.8&46.1&24.1&38.8&41.0\\
Qwen2.5-7B&26.6&9.9&63.5&29.5&27.9&20.1&32.7&33.8&18.0&28.6&32.0\\
Qwen2.5-3B&16.2&12.8&71.0&18.6&17.3&13.4&17.9&26.1&9.3&15.6&20.8\\
Qwen2.5-1.5B&11.1&14.6&74.3&13.1&12.0&11.0&11.3&18.7&6.7&12.2&12.9\\
\midrule
GLM4-9B&25.9 & 12.5 & 61.6 & 29.6 & 27.6 & 28.8 & 32.1 & 32.0 & 17.6 & 28.9 & 27.8   \\ 
ChatGLM3-6B  &11.2&13.6&75.2&12.9&12.0&12.1&13.8&12.4&8.8&13.4&11.8 \\ 
\midrule
InternLM2.5-20B&31.5&7.7&60.8&34.1&32.8&32.0&37.1&37.7&21.2&35.7&34.3\\
InternLM2.5-7B&24.7&7.5&67.8&26.7&25.7&25.5&29.4&31.0&16.4&26.9&25.8\\
InternLM2.5-1.8B&5.3&31.1&63.6&7.6&6.2&6.1&8.7&7.2&3.3&4.5&7.4\\
\midrule
Yi-1.5-34B&30.9&5.8&63.3&32.8&31.8&28.2&36.9&36.8&24.4&32.8&31.4\\
Yi-1.5-9B&18.2&2.9&78.9&18.7&18.4&17.2&20.2&24.3&10.2&20.1&19.8\\
Yi-1.5-6B&15.9&2.8&81.3&16.3&16.1&14.2&17.9&21.3&10.3&16.8&16.5\\
\midrule
LLaMA3.1-70B&38.3 & 9.4 & 52.3 & 42.3 & 40.2 & 22.9 & 47.2 & 49.3 & 34.5 & 49.6 & 40.4\\
LLaMA3.1-8B&16.9 & 8.8 & 74.3 & 18.6 & 17.7 & 8.5 & 20.7 & 23.4 & 9.7 & 20.5 & 20.7\\
\midrule
DeepSeek-67B&43.5 & 14.8 & 41.7 & 51.1 & 47.0 & 34.3 & 54.5 & 50.3 & 42.3 & 49.0 & 46.2	\\
DeepSeek-V2-Lite-Chat &33.7&12.8&53.5&38.6&36.0&35.3&38.5&41.7&32.2&37.5&31.2\\
% DeepSeek-V2	\\
DeepSeek-7B&23.2 & 13.2 & 63.6 & 26.7 & 24.8 & 24.5 & 27.2 & 28.9 & 20.6 & 27.0 & 21.5	\\
\midrule
Baichuan2-13B&19.1&24.9&56.0&25.4&21.8&24.0&25.8&23.3&16.8&23.0&18.7\\
Baichuan2-7B&12.5 & 21.8 & 65.7 & 16.0 & 14.0 & 14.6 & 16.1 & 15.4 & 11.1 & 13.8 & 13.3\\
\midrule
Mixtral-8x22B-Instruct-v0.1&27.3 & 2.2 & 70.5 & 27.9 & 27.6 & 10.6 & 32.3 & 36.0 & 21.0 & 34.1 & 26.9\\
Mixtral-8x7B-Instruct-v0.1&20.4 & 7.2 & 72.4 & 22.0 & 21.2 & 5.2 & 26.5 & 29.0 & 13.0 & 25.0 & 23.3\\
Mistral-7B-Instruct-v0.2&15.0 & 8.8 & 76.2 & 16.4 & 15.6 & 4.5 & 18.2 & 22.2 & 9.5 & 21.4 & 15.7\\
% \midrule
% Gemma2-2B \\
% Gemma2-9B \\
% Gemma2-27B \\

\bottomrule
\end{tabular}
}
% \vspace{-2mm}
% Note that ``ZS'', ``ZS-COT'', ``FS'' represents ``zero-shot'', ``zero-shot w/ chain-of-thought'' and ``few-shot'', repsectively. Models are grouped roughly according to their model sizes.}
% \vspace{-5mm}
\caption{Results of different models on Chinese SimpleQA. For metrics, 
\textbf{CO},   \textbf{NA},    \textbf{IN},    and \textbf{CGA} denote ``Correct'', ``Not attempted'', ``Incorrect'', and ``Correct given attempted'', respectively.
For subtopics, \textbf{CC}, \textbf{HU}, \textbf{ETAS}, \textbf{LAC}, \textbf{SO} and \textbf{NS} represent ``Chinese Culture'', ``Humanities'', ``Engineering, Technology, and Applied Sciences'', ``Life, Art, and Culture'', ``Society'', and ``Natural Science'', respectively. 
% Following SimpleQA~\citep{Wei2024MeasuringSF}, \textbf{F-score} is the harmonic mean between correct and correct given attempted.
}
\label{tab:mainresults}
\end{table*}
 
\subsection{Main Results}
\begin{figure}[t]
\centering
\includegraphics[width=\textwidth]{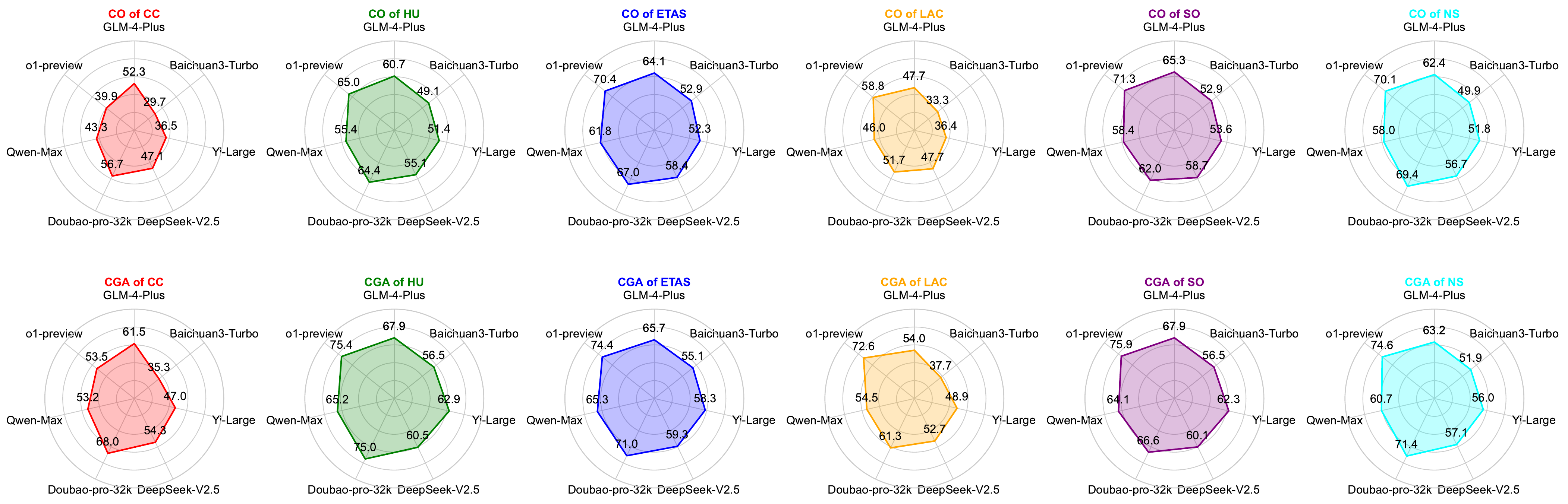}
% \vspace{-0.6cm}
\caption{Results (\textbf{CO} and \textbf{CGA} metrics) of different models for six topics.}
\label{fig:main_fig}
%\vspace{-0.6cm}
\end{figure}

As shown in Table~\ref{tab:mainresults},
we provide the performance results of different LLMs on our  Chinese SimpleQA.
Specifically,
following SimpleQA,
we provide the overall results on 5 evaluation metrics.
Additionally,
we also report the F-score for 6 topics to analyze the fine-grained factuality abilities of these LLMs.
In Table~\ref{tab:mainresults},
we have the following insightful and interesting observations:
\begin{itemize}
\item o1-preview achieves the best performance on Chinese SimpleQA, and the performance results of several recent closed-source LLMs focusing on Chinese (Doubao-pro-32k and GLM-4-Plus) are very close to o1-preview.
\item It is obvious that the ``mini'' series models (o1-mini, GPT-4o-mini) achieve lower results than the corresponding larger models  (o1-preview, GPT-4o),
which also indicates these ``mini'' series models do not pay attention to memorize factuality knowledge.
\item A Larger LLM leads to better performance,
where we can draw this conclusion based on many model series (e.g., GPT, Qwen2.5, InternLM2.5, Yi-1.5)
\item Small LLMs usually lead to higher scores on  ``not attempted (NA)''. The NA scores  for o1-mini, InternLM2.5-1.8B are 20.5 and 31.2,
respectively,
which are larger than the scores of corresponding larger LLMs a lot  (o1-preview with 12.2, InternLM2.5-20B with 7.7).
\item There is a significant performance difference among different subtopics for different LLMs. Notably, the Chinese community LLMs (e.g., Doubao-pro-32k, GLM-4-Plus, Qwen-Max, Deepseek) are significantly better than the GPT or o1 models in the Chinese Culture (CC) subtopic. In contrast, the o1  has significant advantages in science-related subtopics (e.g., Engineering, Technology, and Applied Sciences (ETAS), and Natural Science (NS)).
\end{itemize}

In addition, we also provide the detailed results (CO and CGA metrics) on 6 topics in Figure~\ref{fig:main_fig}. 

\begin{figure}[!t]
\centering
% \setlength{\abovecaptionskip}{0.1cm}
% \setlength{\belowcaptionskip}{-0.3cm}  
% \resizebox{0.48\textwidth}{!}{}
\includegraphics[width=1\linewidth]{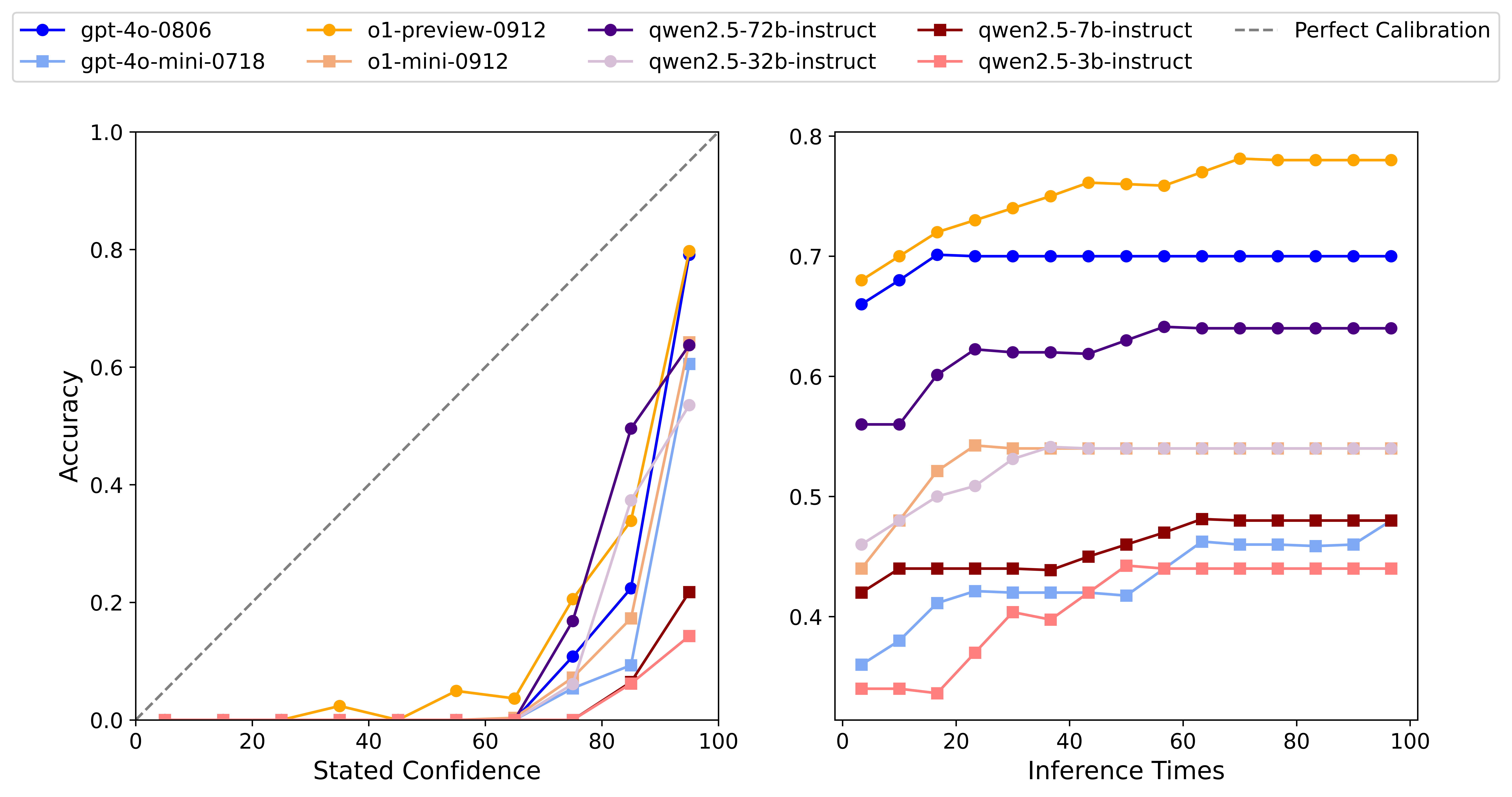}
% \vspace{-0.6cm}
\caption{Left: Calibration of LLMs based on their stated confidence. Right: Improvement in accuracy with increased test-time compute using Best-of-N.}
\label{fig: exp1}
% \vspace{-0.6cm}
\end{figure}
\subsection{Further Analysis}

\subsubsection{Analysis of Calibration}
% We analyzed the calibration of different LLMs on Chinese SimpleQA and the accuracy improvements caused by increased test-time compute. 
For the calibration of different LLMs, 
following SimpleQA, we instruct the model to provide a corresponding confidence level (from 0 to 100) when answering questions to measure the model's confidence in its answers (See the prompt in Appendix \ref{app: calibration}). We know that a perfectly calibrated model's confidence (\%) should match its answers' actual accuracy. The left plot in Figure \ref{fig: exp1} illustrates the alignment performance, 
which indicates that GPT-4o aligns better than GPT-4o-mini and o1-preview aligns better than o1-mini.
For the Qwen2.5 series, the alignment order is Qwen2.5-72B \textgreater{} Qwen2.5-32B \textgreater{} Qwen2.5-7B \textgreater{} Qwen2.5-3B,
which suggests that larger model sizes result in better calibration. 
% Among them, o1-preview has the best alignment, followed by qwen-72b-instruct. The alignment of o1-mini is similar to that of 4o-mini. 
Furthermore, for all evaluated models, their confidence in the range of confidence \textgreater{} 50 falls below the line of perfect alignment, which means that they all overestimate the accuracy of their responses and overconfidence exists.

\subsubsection{Analysis of Test-Time Compute}
We also evaluate the relationship between increased test-time compute and response accuracy for different models. Specifically, we randomly sample 50 samples from Chinese SimpleQA, and for each sample, the model is asked to independently answer 100 times. Then, we obtain the model's response accuracy using the Best-of-N method as the inference counts increase. The results are shown in the right plot of Figure \ref{fig: exp1}. We observe that as the times of inferences increase, the response accuracy of all models improves and eventually reaches a ceiling. This is reasonable for Chinese SimpleQA, which is specifically designed to probe the boundaries of a model's knowledge. 
% For questions where the model has some relevant knowledge but is not entirely certain, it can make repeated attempts until it answers correctly. However, for questions it knows nothing about, increasing test-time compute will not yield correct answers.

% openai两张图，方差柱状图
% \subsubsection{Analysis on the System Prompt}
\begin{wrapfigure}[18]{r}{0.5\textwidth}
    \centering
    \includegraphics[width=0.5\textwidth]{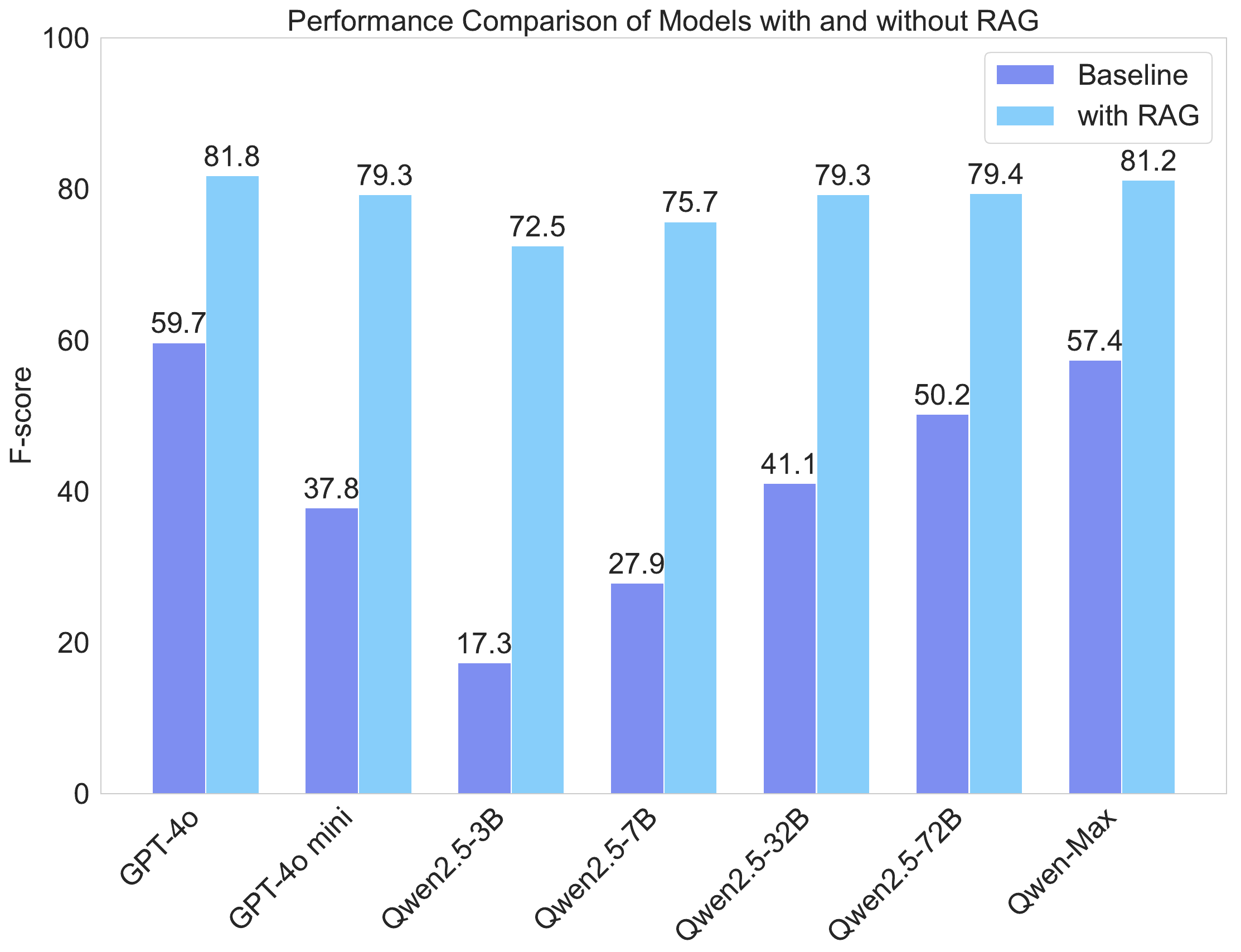}
    \caption{The effect of RAG strategy.}
    \label{fig: rag}
\end{wrapfigure}
\subsubsection{Analysis on the effect of RAG}

% \begin{table}[!t]
% \centering
% \begin{tabular}{cccccc}
% \toprule
% \textbf{Model} & \textbf{CO} & \textbf{NA} & \textbf{IN} & \textbf{CGA} & \textbf{F1-score} \\ \midrule
% \multicolumn{1}{l}{GPT-4o} & 59.3 & 1.3 & 39.3 & 60.1 & 59.7 \\
% w. RAG & 78.4 & 4.2 & 13.2 & 85.6 & 81.8 \\ \midrule
% \multicolumn{1}{l}{GPT-4o mini} & 37.6 & 0.9 & 61.5 & 37.9 & 37.8 \\
% w. RAG & 76.8 & 2.0 & 17.0 & 81.9 & 79.3 \\ \midrule
% \multicolumn{1}{l}{Qwen2.5-3B} & 16.2 & 7.6 & 71.0 & 18.6 & 17.3 \\
% w. RAG & 68.8 & 4.4 & 20.9 & 76.7 & 72.5 \\ \midrule
% \multicolumn{1}{l}{Qwen2.5-7B} & 26.6 & 4.6 & 63.5 & 29.5 & 27.9 \\
% w. RAG & 71.1 & 6.0 & 16.9 & 80.8 & 75.7 \\ \midrule
% \multicolumn{1}{l}{Qwen2.5-32B} & 38.8 & 5.6 & 50.1 & 43.6 & 41.1 \\
% w. RAG & 74.1 & 7.2 & 12.8 & 85.2 & 79.3 \\ \midrule
% \multicolumn{1}{l}{Qwen2.5-72B} & 48.4 & 1.8 & 44.5 & 52.1 & 50.2 \\
% w. RAG & 74.9 & 4.2 & 13.9 & 84.3 & 79.4 \\ \midrule
% \multicolumn{1}{l}{Qwen-Max} & 54.1 & 5.8 & 34.6 & 61.0 & 57.4 \\
% w. RAG & 76.7 & 5.2 & 12.2 & 86.3 & 81.2 \\ \bottomrule
% \end{tabular}
% % \vspace{-0.2cm}
% \caption{temp.}
% % \vspace{-0.5cm}
% \label{tab: RAG}
% \end{table}

% \begin{figure}[!t]
% \centering
% % \setlength{\abovecaptionskip}{0.1cm}
% % \setlength{\belowcaptionskip}{-0.3cm}  
% % \resizebox{0.48\textwidth}{!}{}
% \includegraphics[width=1\linewidth]{figures/exp1.pdf}
% % \vspace{-0.6cm}
% \caption{temp.}
% \label{fig: exp1}
% % \vspace{-0.6cm}
% \end{figure}

In this study, we explore the effectiveness of the Retrieval-Augmented Generation (RAG) strategy in enhancing the factual accuracy of large language models (LLMs) on the Chinese SimpleQA dataset.
Specifically, we reproduce a RAG system based on LlamaIndex~\citep{Liu_LlamaIndex_2022}, incorporating Google search APIs.
As illustrated in Figure~\ref{fig: rag}, all models demonstrate a substantial improvement in accuracy with RAG. For example, the performance of Qwen2.5-3B improved more than threefold. Notably, nearly all models with RAG outperform the native GPT4-o model. 
Meanwhile, the application of RAG also leads to a marked reduction in performance disparities among models. For example, the F-score difference between the Qwen2.5-3B with RAG and the Qwen2.5-72B with RAG is only 6.9\%.
% The application of RAG also leads to a marked reduction in performance disparities among models. For instance, the F-score difference between Qwen2.5-3B with RAG and Qwen2.5-72B with RAG is only 6.9\%. 
This suggests that RAG reduces the performance gaps on models greatly, enabling even smaller ones to achieve high performance when augmented with RAG. Overall, this suggests that RAG serves as an effective shortcut for enhancing the factuality of LLMs.

% 基于llamaindex reproduce了一个RAG系统

% \subsubsection{Analysis on the effect of RAG}

% \subsubsection{Analysis on the ``Know How''}

% \subsubsection{Analysis on the effect of Chain-of-Thought}
% % 若干种cot prompt，验证
% \begin{figure}[!t]
% \centering
% % \setlength{\abovecaptionskip}{0.1cm}
% % \setlength{\belowcaptionskip}{-0.3cm}  
% % \resizebox{0.48\textwidth}{!}{}
% \includegraphics[width=1\linewidth]{figures/exp1.pdf}
% % \vspace{-0.6cm}
% \caption{temp.}
% \label{fig: exp1}
% % \vspace{-0.6cm}
% \end{figure}
% Additionally, we also analyze the effect of the Chain-of-Thought strategy on factuality. We reproduce two classic Chain-of-Thought strategies: "Recall and Answer", and "Step by Step". The results are shown in Figure ~\ref{fig: exp1}(b). It can be seen that the performance changes brought by the two COT strategies are not significant.
% % 【TDO】这块实验要再re一下，现在不好分析， 区分domain来看？

\subsubsection{Analysis on the alignment tax}
\begin{figure}[!t]
\centering
% \setlength{\abovecaptionskip}{0.1cm}
% \setlength{\belowcaptionskip}{-0.3cm}  
% \resizebox{0.48\textwidth}{!}{}
\includegraphics[width=1\linewidth]{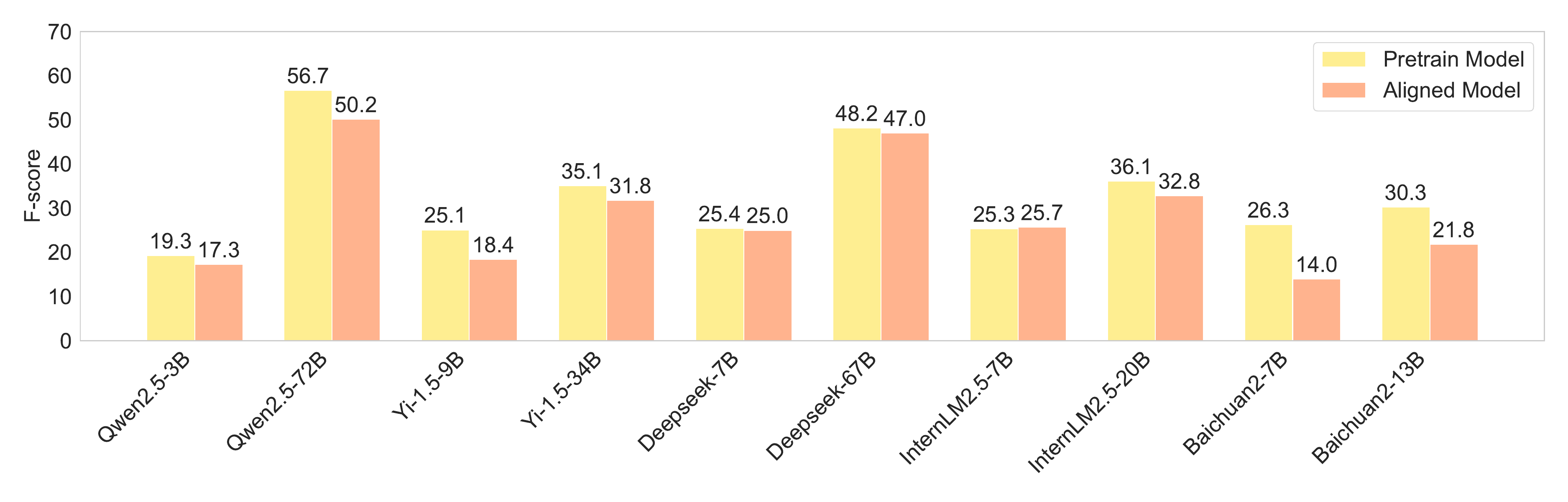}
% \vspace{-0.6cm}
\caption{The effect of alignment in post-training.}
\label{fig: exp3}
% \vspace{-0.6cm}
\end{figure}

% \begin{figure}[!t]
% \centering
% % \setlength{\abovecaptionskip}{0.1cm}
% % \setlength{\belowcaptionskip}{-0.3cm}  
% % \resizebox{0.48\textwidth}{!}{}
% \includegraphics[width=1\linewidth]{figures/exp3_1.pdf}
% % \vspace{-0.6cm}
% \caption{temp.}
% \label{fig: exp3-1}
% % \vspace{-0.6cm}
% \end{figure}
Recently, prior studies~\citep{gpt4,song2023reward} have found that the alignment can lead to a decrease in the abilities of language models as known as the ``alignment tax''.
To illustrate the effect of alignment on factuality, we conduct a comparative performance analysis between pre-trained models and aligned models that are trained with Supervised Fine-Tuning (SFT) or Reinforcement Learning from Human Feedback (RLHF). 
As illustrated in Figure~\ref{fig: exp3}, different models exhibit varying trends after post-training, but most models have a significant decline. Among these, the Baichuan2 series models show the most significant decreases, with Baichuan2-7B and Baichuan2-13B experiencing F-score reductions of 47\% and 28\%, respectively.
% Further analysis reveals that aligned models are more prone to generating knowledge hallucinations, particularly in time-related questions.
This reflects that the alignment training of most current LLMs still has obvious drawbacks to produce knowledge hallucinations, which further reflects the necessity of our dataset.
%the performance of most models decreases after Supervised Fine-Tuning (SFT) or Reinforcement Learning from Human Feedback (RLHF). 
%For example, the Qwen2.5-72B model shows a reduction from an F1-score of 56.7 in its pre-trained state to 50.2 after aligned, and The F-score of Baichuan-7B dropped by nearly 47\%. 

%Among them, the Baichuan2 series models performed most significantly, with Baichuan2-7B decreasing by 84.8\%. Further analysis shows that the aligned model is more likely to produce knowledge hallucinations. This reflects that the alignment training of most current models still has obvious drawbacks, and further reflects the value of our dataset.

\subsubsection{Analysis on the results of subtopics}
% 再加个回答类型
\begin{figure}[!t]
\centering
% \setlength{\abovecaptionskip}{0.1cm}
% \setlength{\belowcaptionskip}{-0.3cm}  
% \resizebox{0.48\textwidth}{!}{}
\includegraphics[width=1\linewidth]{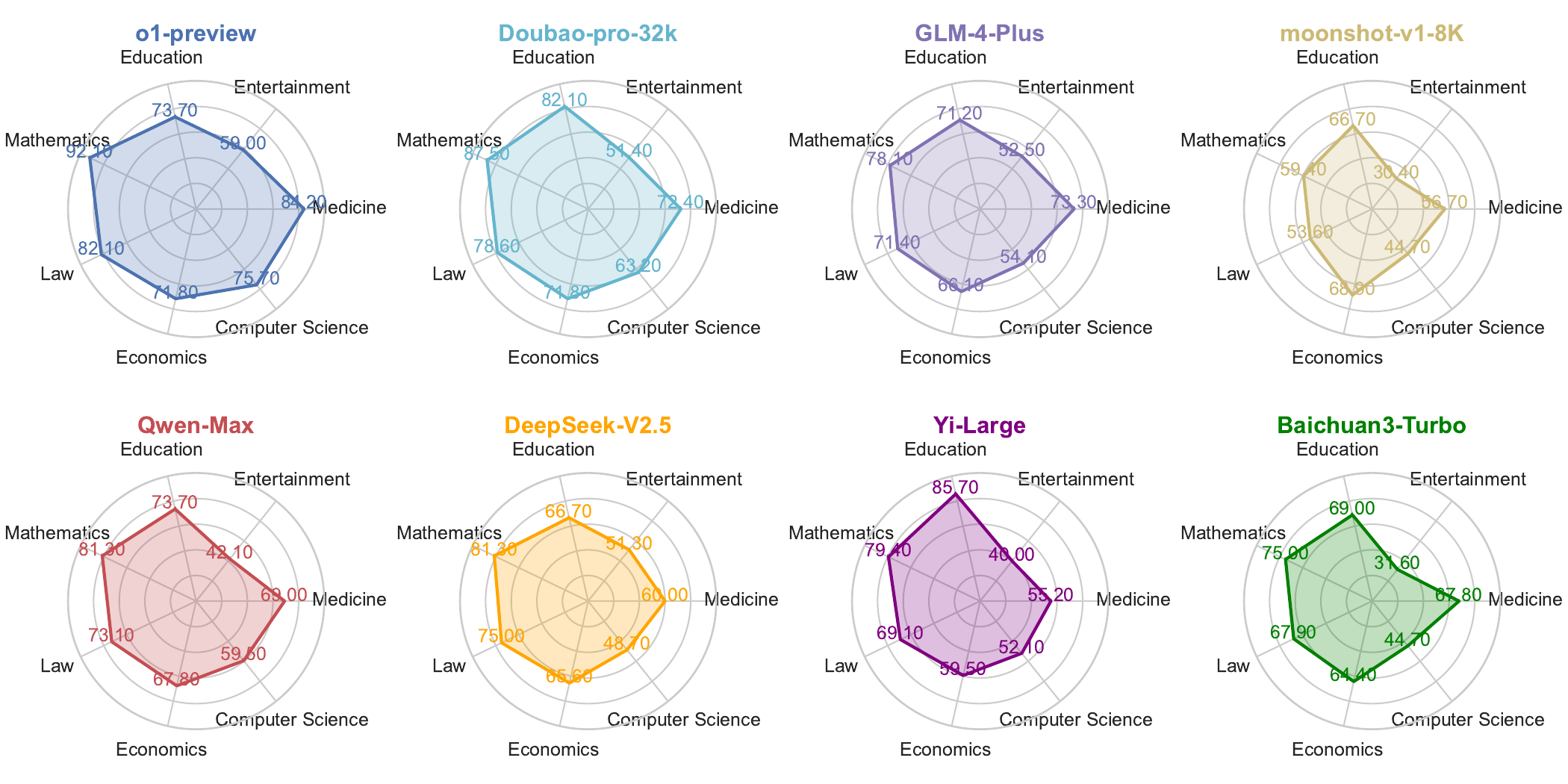}
% \vspace{-0.6cm}
\caption{Detailed results on some selected subtopics.}
\label{fig: exp4}
% \vspace{-0.6cm}
\end{figure}
As mentioned in Section~\ref{Data collection}, the benchmark covers a total of 99 subtopics, which can comprehensively detect the knowledge level of the model in various domains. Figure~\ref{fig: exp4} illustrates the performance comparison between the o1 model and seven notable Chinese community models within several common domains. 
Firstly, from an overall perspective, the o1-preview model exhibits the most comprehensive performance across these domains, with the Doubao model following closely. In contrast, the Moonshot model demonstrates the weakest overall performance.
Secondly, when examining specific domains, a significant disparity emerges between the Chinese community models and the o1 model in areas such as Computer Science and Medicine. However, this gap is minimal in domains like Education and Economics. Notably, in Education, some Chinese community models outperform the o1-preview, highlighting their potential for achieving success in specific vertical domains.
Lastly, when examining specific models, the Moonshot model is notably weaker in Mathematics, Law, and Entertainment, while the Baichuan model also underperforms in Entertainment. The Yi-Large model excels in Education, and the o1 model maintains the strongest performance across other domains. 
Evaluating the performance of the models across diverse domains within the benchmark dataset enables users to identify the most suitable model for their specific needs.

\subsubsection{Comparison between Chinese SimpleQA and SimpleQA}
% 柱状图分析
\begin{figure}[!t]
\centering
% \setlength{\abovecaptionskip}{0.1cm}
% \setlength{\belowcaptionskip}{-0.3cm}  
% \resizebox{0.48\textwidth}{!}{}
\includegraphics[width=1\linewidth]{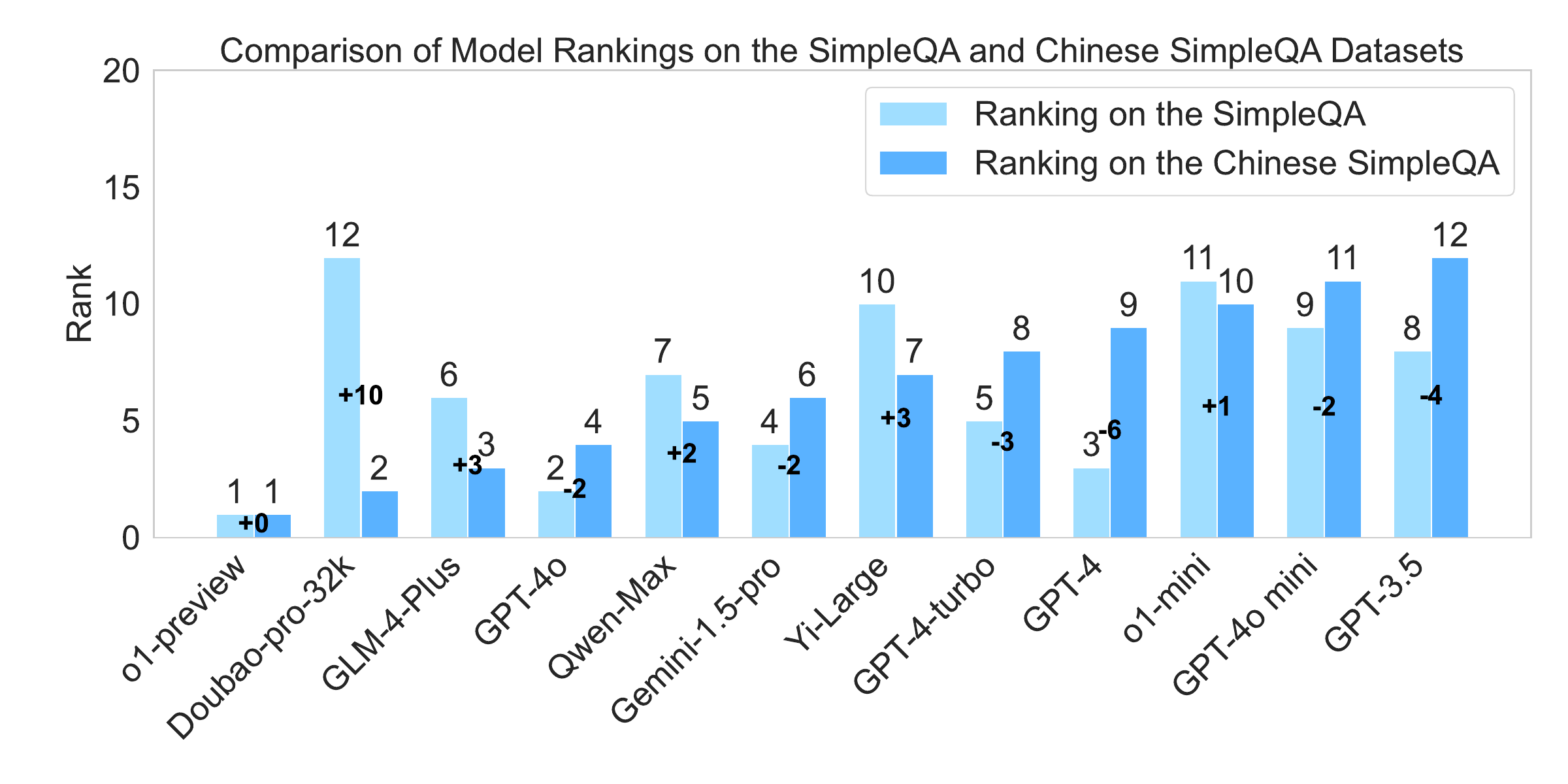}
\vspace{-4mm}
\caption{The rankings of different LLMs on SimpleQA and Chinese SimpleQA.}
\label{fig: exp6}
% \vspace{-0.6cm}
\end{figure}

We also compare the ranking differences of various models on the SimpleQA and the Chinese SimpleQA. As illustrated in Figure~\ref{fig: exp6}, there are notable discrepancies in model performance across these two benchmarks. For instance, Doubao-pro-32k ranks significantly higher on the Chinese SimpleQA, moving from 12th to 2nd place (+10). Conversely, GPT-4 shows a decline in performance on the Chinese SimpleQA, dropping from 3rd to 9th place (-6).
These differences emphasize the importance of evaluating models on datasets in various languages and the need for research into optimizing model performance across different linguistic environments.
Notably, the o1-preview maintains its top position consistently across both datasets, indicating its robustness and adaptability to different linguistic contexts. 
In addition, most Chinese community-developed models (e.g., Qwen-Max, GLM-4-Plus, Yi-Large, Doubao-pro-32k) perform better on the Chinese SimpleQA than on the SimpleQA, demonstrating their competitive results on Chinese-language tasks.

\section{Related Works}

% The concept of LLM factuality pertains to the ability of large language models to generate content that adheres to verifiable information, encompassing general knowledge, domain-specific facts, and common sense. This factual content can be substantiated by authoritative sources such as encyclopedias, specialized textbooks, and reputable dictionaries.

% Researchers have explored the potential of LLMs to function as repositories of factual knowledge [1,2,3]. Existing studies have primarily focused on qualitative assessments of LLM factuality [4,5], investigations into knowledge storage mechanisms [6,7], and analyses of the origins of knowledge-related issues [8,9]. Among these areas, the question of factuality in LLMs has garnered significant attention.

% LLMs may exhibit deficiencies in specialized knowledge domains, such as medicine or law. Moreover, they may lack awareness of events occurring after their most recent training update. In some instances, despite possessing relevant information, LLMs may struggle to deduce correct answers through logical reasoning. Additionally, these models might experience difficulty recalling previously acquired knowledge.

% The factuality challenge intersects with several prominent research areas in the field of Large Language Models, including:

% Hallucinations: The generation of plausible but fictitious information [10]
% Outdated Information: The reliance on obsolete or superseded facts [11]
% Domain-Specificity: The application of LLMs to specialized fields such as law [12] and finance [13]

\noindent\textbf{LLM Factuality.}
LLM factuality is the capability of large language models to produce contents that follow factual content, including commonsense, world knowledge, and domain facts,
and the factual content can be substantiated by authoritative sources (e.g., Wikipedia, textbooks).
% The factual information can be grounded to reliable sources, such as dictionaries, Wikipedia or textbooks from different domains.
Recent works have explored the potential of LLMs to serve as factual knowledge bases~\citep{yu2023generate,pan2023unifying}.
Specifically,
existing studies have primarily focused on qualitative assessments of LLM factuality~\citep{TruthfulQA,chern2023factool}, investigations into knowledge storage mechanisms \citep{meng2022locating,chen2023journey}, and analyses on knowledge-related issues~\citep{gou2023critic}. 
% Among these areas, the question of factuality in LLMs has garnered significant attention.
% Existing works focus on measuring factuality in LLMs qualitatively \citep{TruthfulQA,chern2023factool}, discussing the mechanism for storing knowledge \citep{meng2022locating,chen2023journey} and tracing the source of knowledge issues \citep{gou2023critic,kandpal2023large}. The factuality issue for LLMs receive relatively the most attention.
% For instance, an LLM might be deficient in domain-specific factual knowledge, such as medicine or law domain. Additionally, the LLM might be unaware of facts that occurred post its last update. There are also instances where the LLM, despite possessing the relevant facts, fails to reason out the correct answer. In some cases, it might even forget or be unable to recall facts it has previously learned.
% The factuality problem is closely related to several hot topics in the field of Large Language Models, including {Hallucinations} \citep{Hallucination_Survey}, {Outdated Information} \citep{WebGPT}, and {Domain-Specificity} (e.g., Law \citep{ChatLaw}, Finance \citep{BloombergGPT}). 

\noindent\textbf{Factuality Benchmarks.}
% There are many 
Many factuality benchmarks~\citep{mmlu,zhong2023agieval,huang2023ceval,li2023cmmlu,BigBench,hotpotqa} have been proposed.
For example,
MMLU ~\cite{mmlu} is to measure the multitask accuracies on a diverse set of 57 tasks.
TruthfulQA~\citep{TruthfulQA} focuses on assessing the truthfulness of a language model's generated answers. 
Additionally,
HaluEval \citep{li2023halueval}
is to examine the tendency of LLMs to produce hallucinations.
% where hallucination denotes that the content either conflicts with the source or cannot be verified using factual knowledge. 
Recently, SimpleQA~\citep{Wei2024MeasuringSF} has been proposed to measure the short-form factuality in LLMs.
However, SimpleQA only focuses on the English domain. In contrast, our Chinese SimpleQA aims to evaluate factuality in Chinese comprehensively.

\section{Conclusion}
In this paper, to evaluate the factuality abilities of existing LLMs, we propose the first Chinese short-form factuality benchmark (i.e., Chinese SimpleQA), which
% which includes 6 main topics and 99 subtopics.
% Besides,
mainly has five important features (i.e., Chinese, diverse, high-quality, static, and easy-to-evaluate).
Based on Chinese SimpleQA, we comprehensively evaluate the performance of existing 40+ LLMs on factuality and provide detailed analysis to demonstrate the advantage and necessity of our Chinese SimpleQA.
In the future, we will investigate how to improve the LLMs' factuality and explore how to extend Chinese SimpleQA to multilingual, multimodal, and domain-specific (e.g., code, e-commerce) settings.
\clearpage
\bibliography{iclr2025_conference}
\bibliographystyle{iclr2025_conference}

\clearpage

\appendix
\section{Generation and validation of question-answer pairs}
\label{app: gereration_and_validation}
The generation and validation of question-answer pairs both use OpenAI's gpt-4o-0806. The specific prompts are shown in Figures \ref{fig: gen_qa_pair}, \ref{fig: validate_criteria}, and \ref{fig: verify_factual}.

\begin{figure}[htbp]
\begin{tcolorbox}
\begin{CJK}{UTF8}{gbsn}
现在需要你根据给定的文档生成一个事实类问题和对应的标准答案，需要满足下列要求:\\
1. 生成的问题必须关联到客观世界的知识，例如可以询问“2024年诺贝尔物理学奖的获得者是谁？”不得构造涉及个人观点或感受相关的主观问题，如“你如何看待xxx？”。\\
2. 所提出的问题应该有且只有一个明确且无争议的实体作为答案，且问题表述中不应存在任何形式的模糊性或歧义。例如，避免提问“巴拉克和米歇尔·奥巴马在哪里会面？”因为无法确定是指哪一次会面；同样不要问“白民国人身体的特点是什么？”因为这个问题过于模糊，没有明确的答案。“周汝昌最为人熟知的著作是哪个？”也是不合格问题，因为“最熟知”可能是有争议的。\\
3. 问题的答案应当是时间不变的，不会随着时间的推移而改变。例如，“美国现任总统是谁？”就不是一个合适的问题，因为总统身份会随选举结果改变。\\
4. 问题应该具有一定的难度，以体现出一定的挑战性。例如：电影《脱衣舞娘》是由同名小说改编的，该小说的作者是谁？\\
5. 如果问题的答案为英文人名，请给出中文翻译后的名字和括号里带上英文原名，格式如：雅各布·福格（Jakob Fugger）。\\
6. 生成的问题需要与给定的类目相关\\
请将生成的问题和答案以JSON格式返回，具体格式如下：\\
\{"question": "这里填写生成的问题", "answer": "这里填写对应的标准答案"\}\\
\#\#\#以下是一些示例\#\#\#\\
\#\#\# 示例一\\
类目：娱乐\\
文档内容：2022年国际足联世界杯为第22届国际足联世界杯，于2022年11月20日至12月18日在卡塔尔举行[2][3]，成为全球爆发防疫后首个终结限制的大型国际体育盛事。考量到气候因素，本届世界杯亦是首次于11月至12月北半球秋季[注 1]举行之世界杯。决赛于卡塔尔的卢赛尔体育场举行，由阿根廷队对阵卫冕冠军法国队。双方先于比赛正规时间踢至加时赛以3–3赛和，后在点球大战中阿根廷以4–2击败法国，赢得了此届世界杯，也是阿根廷继1986年世界杯后，相隔36年再度于世界杯夺冠，继巴西，意大利及德国后第四支三次冠军的球队，也成为继巴西后第二支在亚洲夺冠的南美洲球队。克罗地亚则以2比1击败该年黑马摩洛哥赢得季军[4][5]。\\
返回结果：\{"question": "2022年世界杯决赛点球大战中阿根廷队是以多少击败法国队", "answer": "4–2"\}\\
\#\#\# 示例二\\
类目：政治\\
文档内容：中国与世界贸易组织（英语：China and the World Trade Organization）指中华人民共和国与世界贸易组织的关系。在部长级会议达成协议后，中国于2001年12月11日成为世界贸易组织成员。[1][2]在承认这一点之前，双方进行了漫长的谈判，并且需要对中国经济进行重大改革。世贸组织的成员资格一直存在争议，对其它国家产生了重大的经济和政治影响(也被称为“中国冲击”) ，对世贸组织框架与中国经济模式之间的不匹配也存在争议。[3][4]评估和执行合规已成为中美贸易关系中的问题，包括中国的不合规行为如何为本国经济创造利益。[5][6]\\
返回结果：\{"question": "中国是哪一年正式成为世界贸易组织成员？", "answer": "2001"\}\\
\#\#\# \\
让我们开始吧！
\end{CJK}
\end{tcolorbox}
\caption{The prompt for generating question-answer pairs.}
\label{fig: gen_qa_pair}
\end{figure}

\begin{figure}[htbp]
\begin{tcolorbox}
\begin{CJK}{UTF8}{gbsn}
你是一个数据质量检查员，现在需要你检查下面生成的问题是否满足以下要求：\\
1.生成的问题必须对客观世界的知识的提问，例如可以询问“2024年诺贝尔物理学奖的获得者是谁？”不得构造涉及个人观点或感受相关的主观问题，如“你如何看待xxx？”。\\
2. 问题应该有且只有一个明确且无争议的实体作为答案，且问题表述中不应存在任何形式的模糊性或歧义。例如，避免提问“巴拉克和米歇尔·奥巴马在哪里会面？”因为无法确定是指哪一次会面；同样不要问“白民国人身体的特点是什么？”因为这个问题过于模糊，没有明确的答案。注意如果回答是多个实体也不满足要求，例如：“软体动物、腕足动物及被囊动物”\\
3. 问题的答案应当是时间不变的，不会随着时间的推移而改变。例如，“美国现任总统是谁？”就不是一个合适的问题，因为总统身份会随选举结果改变。\\
如果问题不合格则解释并输出“【否】”， 如果问题合格则直接输出“【是】”\\
\#\#\#\# 以下是一些示例：\\
问题：《黄帝内经》中, 援物比类思维方式包括哪些核心概念?\\
评价：该问题不是只有一个确切答案，【否】\\
\\
问题：建筑理论主要研究什么内容？\\
评价：该问题不具体，回答不是只有一个实体，【否】\\
\\
问题：成立了程派高氏八卦掌的高义盛原籍是哪里？\\
评价：回答范围不明确，不清楚是回答到城市还是省份，【否】\\
\\
问题：自由恋爱主义最初的目标是将哪些事务与国家分离？\\
评价：该问题不是只有一个答案，【否】\\
\\
问题：汉十高速公路连接的武汉市和哪两个城市？\\
评价：【是】\\
\#\#\#\#
如果问题不合格则输出原因并最后输出“【否】”， 如果问题合格则直接输出“【是】”，注意如果认为问题不合格需要输出原因\\
让我们开始吧！
\end{CJK}
\end{tcolorbox}
\caption{The prompt for validating criteria.}
\label{fig: validate_criteria}
\end{figure}

\begin{figure}[htbp]
\begin{tcolorbox}
\begin{CJK}{UTF8}{gbsn}
请结合检索材料判断给定的问题是否满足以下要求以及答案是否正确：\\
1. 问题应该有且只有一个明确且无争议的实体作为答案，且问题表述中不应存在任何形式的模糊性或歧义。例如，避免提问“巴拉克和米歇尔·奥巴马在哪里会面？”因为无法确定是指哪一次会面；同样不要问“白民国人身体的特点是什么？”因为这个问题过于模糊，没有明确的答案。注意如果回答是多个实体也不满足要求，例如：“软体动物、腕足动物及被囊动物”\\
2. 问题的答案应当是时间不变的，不会随着时间的推移而改变。例如，“美国现任总统是谁？”就不是一个合适的问题，因为总统身份会随选举结果改变。\\
3. 检索材料应该能够验证给定答案的正确性，并且这些材料中不应存在其他与之相矛盾的答案。如果检索材料中存在不同的答案，则该问题被视为无法验证或答案不明确，不满足要求。\\
请依据上述标准，审查并确保提出的问题及其答案符合要求，如果不满足上面任一要求，则输出原因，最后输出“【不合格】”， 否则输出检索材料对应的片段，最后输出“【合格】”\\
\end{CJK}
\end{tcolorbox}
\caption{The prompt for verifying factual accuracy based on RAG.}
\label{fig: verify_factual}
\end{figure}

\section{Analysis of Model Calibration}
\label{app: calibration}

\begin{figure}[htbp]
\begin{tcolorbox}
\begin{CJK}{UTF8}{gbsn}
请阅读以下问题：\\
\{question\}\\
请基于此问题提供你的最佳答案，并用0到100的分数表示你对该答案的信心（置信度）。请以如下的JSON格式给出回复：\\
\{\\
\hspace*{1em}"answer": "你的答案",\\
\hspace*{1em}"confidence\_score": 你的置信度\\
\}
\end{CJK}
\end{tcolorbox}
\caption{The prompt for guiding the model to output confidence.}
\label{fig: confidence}
\end{figure}

% \clearpage
% \subsection{Error Case of the Step-wise BoN}
% \label{App: Bad Case of the Step-wise BoN}
% \begin{figure}[t]
%     \centering
%     \includegraphics[width=0.9\linewidth]{figure/step_wise.pdf}
%     \caption{The bad case of Step-wise BoN. }
%     \label{fig:step_wise BoN}
% \end{figure}
% As shown in Fig.~\ref{fig:step_wise BoN}, we present the error case of the Step-wise BoN.

% \clearpage
% \subsection{The sample in LiveCodeBench}
% \label{App: The sample in LiveCodeBench}
% \begin{figure}[t]
%     \centering
%     \includegraphics[width=0.99\linewidth]{figure/case_LiveCodeBenchInput.pdf}
%     \caption{The sample in LiveCodeBench. }
%     \label{fig:case_LiveCodeBenchInput}
% \end{figure}

% As shown in Fig~\ref{fig:case_LiveCodeBenchInput}, we present the sample in LiveCodeBench.

\end{document}